\documentclass[conference]{IEEEtran}
\IEEEoverridecommandlockouts
\usepackage{cite}
\usepackage{amsmath,amssymb,amsfonts}
\usepackage{algorithmic}
\usepackage{graphicx}
\usepackage{textcomp}
\def\BibTeX{{\rm B\kern-.05em{\sc i\kern-.025em b}\kern-.08em
    T\kern-.1667em\lower.7ex\hbox{E}\kern-.125emX}}

\usepackage{algorithm}
\usepackage{algorithmic}

\usepackage{amsmath, amsfonts}
\usepackage{multirow, multicol, makecell} 
\usepackage{color, colortbl} 
\usepackage[dvipsnames]{xcolor}

\definecolor{ztfgray}{rgb}{0.85, 0.85, 0.85} 
\newcommand{\CC}[1]{\cellcolor{ztfgray!#1}}

\newcommand{\Reals}{\mathbb{R}}
\newcommand{\dspace}{N}
\newcommand{\dreducedspace}{M}
\newcommand{\trainnumber}{\ndata}
\newcommand{\originalsignal}{\bar{\textbf{x}}}
\newcommand{\optsignal}{\textbf{x}}
\newcommand{\observedsignal}{\textbf{y}}
\newcommand{\measurementmatrix}{\boldsymbol{\Phi}}

\newcommand{\regparamsparse}{\lambda}
\newcommand{\regfunctionsparse}{\mathcal{G}}

\newcommand{\ndata}{\ell}

\begin{document}

\title{LR-CSNet: Low-Rank Deep Unfolding Network for Image Compressive Sensing\\
\thanks{
Corresponding author: Zhenming Peng \\
This work was supported by supported by Natural Science Foundation of Sichuan Province of China (Grant No.2022NSFSC40574)and partially National Natural Science Foundation of China (Grant No.61775030, Grant No.61571096).
}
}

\author{
1$^{\text{st}}$ Tianfang Zhang$^{1,2}$, 
2$^{\text{nd}}$ Lei Li$^{2}$, 
3$^{\text{rd}}$ Christian Igel$^{2}$, 
4$^{\text{th}}$ Stefan Oehmcke$^{2}$, 
5$^{\text{th}}$ Fabian Gieseke$^{2}$, 
6$^{\text{th}}$ Zhenming Peng$^{1,*}$ \\

$^1$School of Information and Communication Engineering, University of Electronic Science and Technology of China
\and 
$^2$Computer Science, University of Copenhagen 
\\
\texttt{sparkcarleton@gmail.com, lilei@di.ku.dk, igel@di.ku.dk} 
\and
\texttt{stefan.oehmcke@di.ku.dk, fabian.gieseke@di.ku.dk, zmpeng@uestc.edu.cn}


}

\maketitle

\begin{abstract}
Deep unfolding networks~(DUNs) have proven to be a viable approach to compressive sensing~(CS). 
In this work, we propose a DUN called low-rank CS network~(LR-CSNet) for natural image CS. 
Real-world image patches are often well-represented by low-rank approximations. 
LR-CSNet exploits this property by adding a low-rank prior to the CS optimization task.
We derive a corresponding iterative optimization procedure using variable splitting, which is then translated to a new DUN architecture.
The architecture uses low-rank generation modules~(LRGMs), which learn low-rank matrix factorizations, as well as gradient descent and proximal mappings~(GDPMs), which are proposed to extract high-frequency features to refine image details.
In addition, the deep features generated at each reconstruction stage in the DUN are transferred between stages to boost the performance.
Our extensive experiments on three widely considered datasets demonstrate the promising performance of LR-CSNet compared to state-of-the-art methods in natural image CS.
\end{abstract}

\begin{IEEEkeywords}
Image Compressive Sensing, Deep Learning, Deep Unfolding Network, Low-Rank Prior, Image Restoration
\end{IEEEkeywords}

\section{Introduction}


Compressive sensing~(CS) has become an important tool in modern signal processing.
It allows to identify  sparse solutions of underdetermined linear systems~\cite{donoho2006compressed}. Under the assumption that the original signal is sparse in some  transform domain~\cite{mallat1999wavelet}, CS requires fewer measurements to reconstruct the original signal than expected by the Nyquist sampling theorem~\cite{donoho2006compressed}. Compressive sensing methods have successfully been applied  in various fields, including single-pixel cameras~\cite{duarte2008single}, magnetic resonance imaging~\cite{lustig2007sparse}, and seismic imaging~\cite{hennenfent2008simply}.

\begin{figure}[t]
  \centering
  \includegraphics[width=0.52\textwidth]{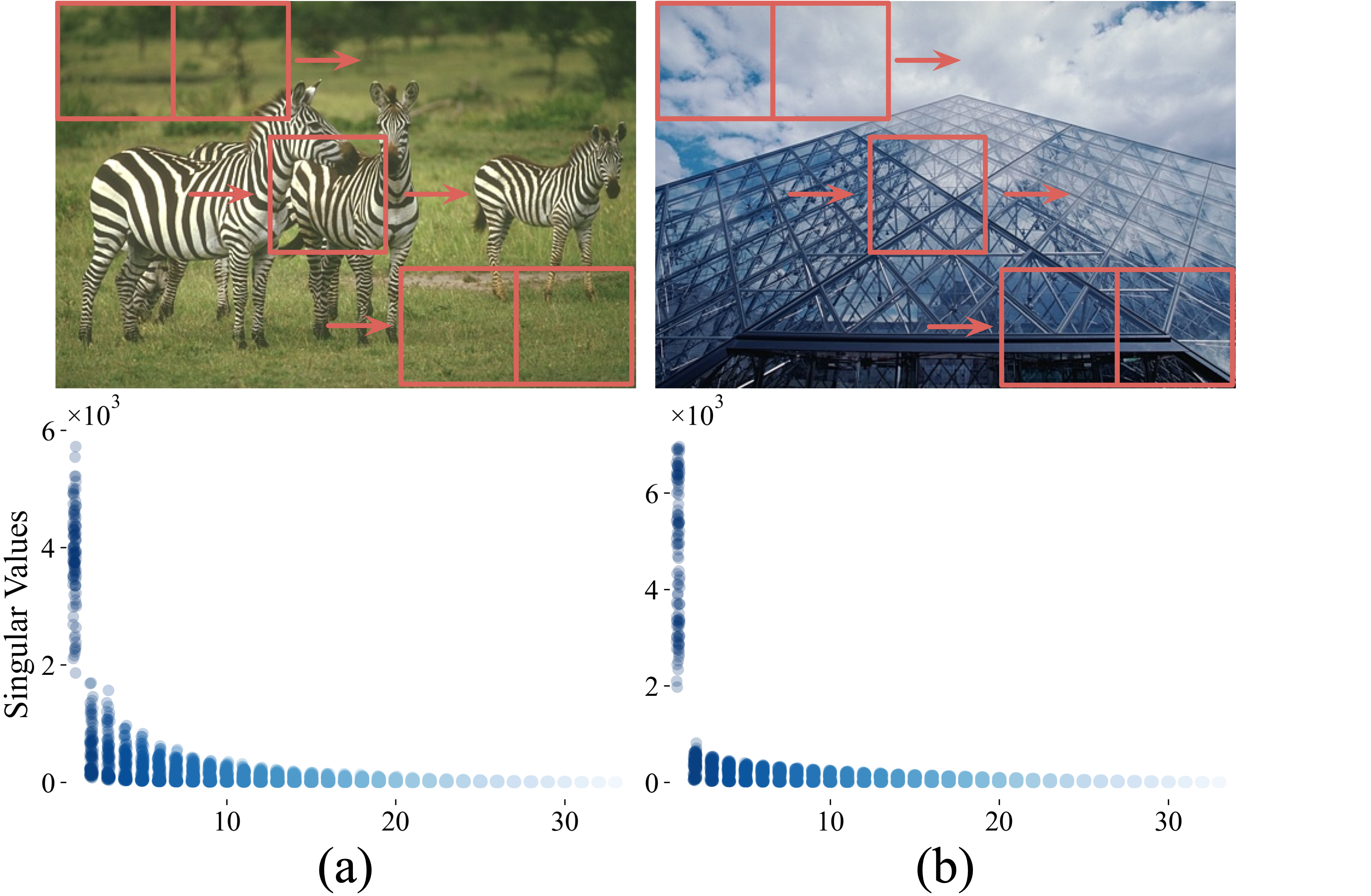}
  \caption{Illustration of the low-rank prior in the images patches. As DUNs operate on image patches, large images are divided into multiple image patches of size 33$\times$33 (red boxes). A singular value decomposition~(SVD) is then performed on all patches per image and the corresponding singular values are plotted into the second row. It can be clearly seen that the singular values quickly decrease, which indicates the low-rank nature of the images.
  }
  \label{fig:lowrankprior}
\end{figure}


Mathematically, for an original signal $\originalsignal \in \Reals^\dspace$, the observation $\observedsignal = \measurementmatrix \originalsignal \in \Reals^\dreducedspace$ is obtained after sampling through a measurement matrix $\measurementmatrix \in \Reals^{\dreducedspace \times \dspace}$, where $\dreducedspace \ll \dspace$.\footnote{Note that for $\originalsignal \in \Reals^{\sqrt{\dspace}\times\sqrt{\dspace}}$ and $\operatorname{vec}(\originalsignal)\in \Reals^\dspace$ being the vectorization of $\originalsignal$ we have $\|\originalsignal\|_{\text{F}} =\|\operatorname{vec}(\originalsignal )\|_2$.} Here,~$\frac{\dreducedspace}{\dspace}$ is denoted as the so-called CS ratio. Given a matrix $\measurementmatrix$ and $\observedsignal$, the goal of compressed sensing is to reconstruct $\originalsignal$ under some sparseness assumptions, such as structural sparsity~\cite{usman2011k}, dictionary sparsity~\cite{ravishankar2010mr}, and low-rankness~\cite{ravishankar2017low}.
We can write CS as an optimization problem of the form
\begin{equation}
    \hat{\textbf{x}} = 
    \arg \min \limits_{\optsignal \in \Reals^\dspace} 
    \left[ 
    \frac{1}{2} \left\| \measurementmatrix \optsignal - \observedsignal \right\|^2_{\text{F}} + \regparamsparse \regfunctionsparse(\optsignal)
    \right]
    \label{eq:csdefinition}
\end{equation}
for $\observedsignal = \measurementmatrix \originalsignal$, where $\left\| \cdot \right\|_{\text{F}}$ is the Frobenius-norm, $\regfunctionsparse$ a sparseness constraint function, and $\regparamsparse$ controls the penalty strength.


In recent years, with the advancement of neural networks, data-driven CS reconstruction methods have made great progress. In general, they can be divided into two categories: deep non-unfolding networks~(DNUNs) and deep unfolding networks~(DUNs). DNUNs learn the mapping between the observed signal $\observedsignal$ and the original signal $\originalsignal$ directly from training examples~\cite{shi2019image}.
In contrast, DUNs consider the optimization problem given by Eq.~\eqref{eq:csdefinition} and map the iterative optimization algorithm used to solve Eq.~\eqref{eq:csdefinition}
to a deep neural network architecture. Usually, $K$ optimization steps are mimicked by means of $K$ sequential blocks (reconstruction stages) in the network~\cite{zhang2020amp}. 
DUNs learn the  matrix $\measurementmatrix$, the regularization function $\regfunctionsparse$, and parameters of the underlying optimization process simultaneously end-to-end by minimizing an objective function of the form 
   
\begin{equation}
   \frac{1}{{2\ndata}}\sum^{{\ndata}}_{i=1} \left\| \originalsignal_i - f_{\text{DUN}}(\measurementmatrix \originalsignal_i)\right\|_{\text{F}}^2 \enspace,
    \label{eq:originproblem_2}
\end{equation}
where $f_{\text{DUN}}$ denotes the neural network and $\left\{ \originalsignal_i \right\}^{\trainnumber}_{i=1}$ are~$\trainnumber$ training instances.
The architecture of $f_{\text{DUN}}(\originalsignal_i)$ results from unfolding the iterative optimization of 
Eq.~\eqref{eq:csdefinition}, the parameters of the network that encode (among others) $\measurementmatrix^{\text{T}}$ as well as~$\regfunctionsparse$ (in our case, $\nabla\regfunctionsparse$). Thus, in contrast to model-based CS, the measurement matrix as well as the sparsity regularization are not given \emph{a priori} but are learned from data.
Because of their excellent reconstruction performance~\cite{zhang2018ista, zhang2020optimization}, DUNs have become state of the art in image CS.

However, DUNs usually constrain the signals $\optsignal$ to be sparse in some transform domain, ignoring other intrinsic properties, such as low-rankness. The manipulation of image patches in CS has become a common practice making the low-rank property more prominent. The singular value curves of several image patches are shown in Fig.~\ref{fig:lowrankprior}. The trend indicated in the graph and the convergence of the singular values to $0$ indicate the low-rankness property of the images at hand. We extend Eq.~\eqref{eq:csdefinition}  by an additional term that reflects the low-rankness of the input signals, i.e., 
\begin{equation}
    \hat{\textbf{x}} = \arg \min \limits_{\textbf{x} \in \mathbb{R}^N} \left[ \frac{1}{2} \left\| \boldsymbol{\Phi} \textbf{x} - \textbf{y} \right\|^2_{\text{F}} + \lambda \mathcal{G}(\textbf{x}) + \mu \mathcal{R}(\textbf{x}) \right] \enspace,
    \label{eq:lrdefinition}
\end{equation}
where $\mathcal{R}(\textbf{x})$ is the function increasing in the rank of the signal $\textbf{x}$  and $\mu$ controls the penalty degree. 

In this paper, we propose an optimization-based deep unfolding network for image compressive sensing, dubbed LR-CSNet, by exploring the low-rank prior of the input images from the perspective of neural networks. 
Our main contributions can be summarized as follows:

\begin{enumerate}
    \item For the problem formulation, we establish an achievable constraint on the low-rank component and demonstrate its iterative optimization process by variable splitting.
    \item We propose LR-CSNet to simulate the iterative optimization process into multiple reconstruction stages and learn an end-to-end mapping between observations and original signal.
    \item We design a low-rank generation modul (LRGM) to learn the low-rank components as well as gradient descent and proximal mapping (GDPM) to refine details of the reconstructed image. Furthermore, we enhance the network representation by feature transmission.
    \item We demonstrate via extensive experiments that LR-CSNet exhibits a superior performance on natural image datasets compared to state-of-the-art approaches. 
\end{enumerate}

\begin{figure*}[t]
  \centering
  \includegraphics[width=0.95\textwidth]{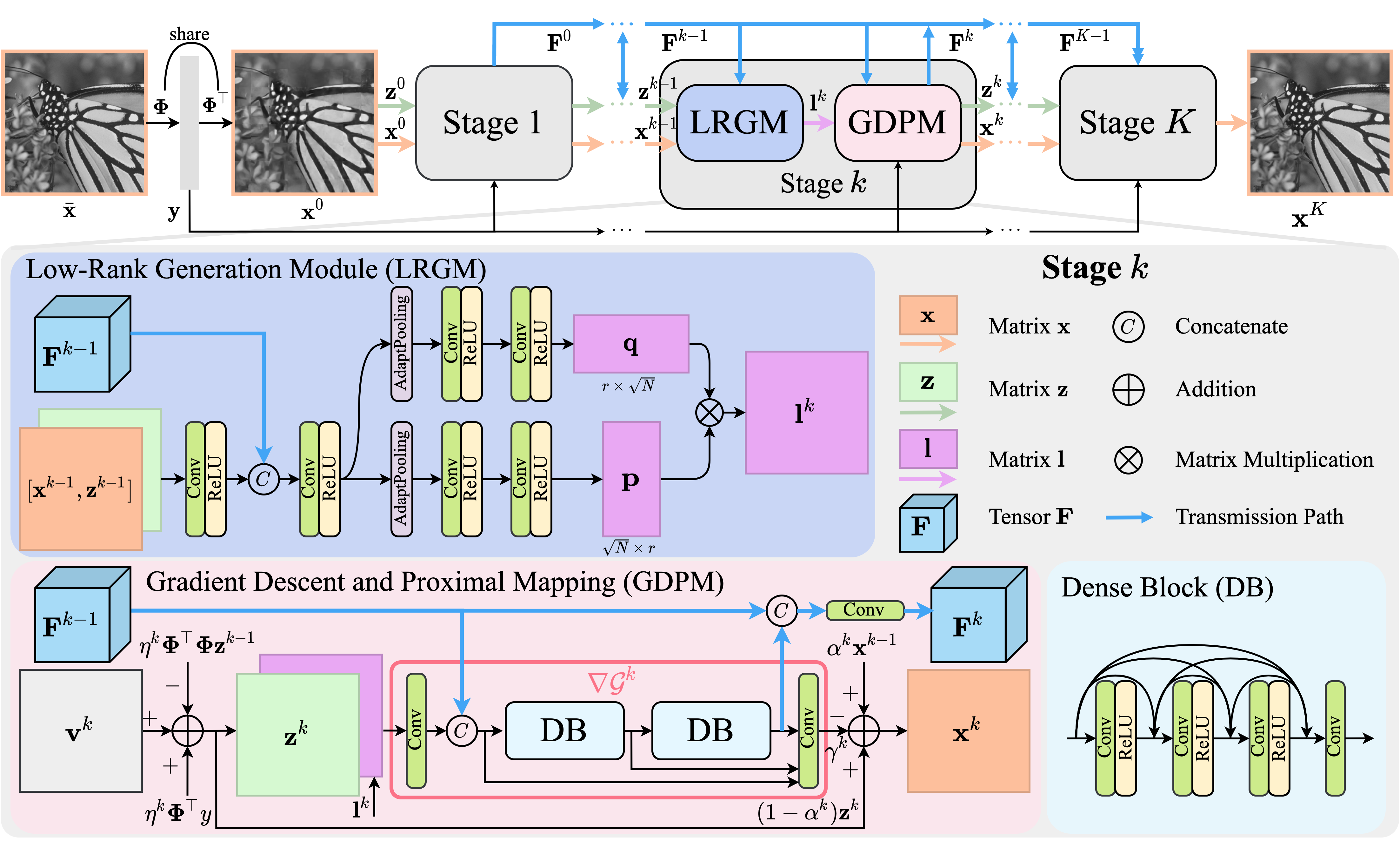}
  \caption{Illustration of the LR-CSNet network architecture and modules, which includes $K$ reconstruction stages in total. From top to bottom is the overall network architecture, LRGM as well as GDPM in the $k$th reconstruction stage, with the legend and dense block~(DB) in GDPM on the right. Specifically, $\textbf{v}^k = \rho^k_1 \textbf{x}^{k-1} + \rho^k_2 \textbf{z}^{k-1} + \left( 1 - \rho^k_1 - \rho^k_2 \right) \textbf{l}^{k}$. $+$ and $-$ indicates the positive and negative sign of the branch when adding up.}
  \label{fig:networkarch}
\end{figure*}

\section{Related Work}

\subsection{Deep Unfolding Networks}

DUNs emulate iterative optimization methods through neural networks and have been successfully applied to image inverse problems~\cite{liu2021stochastic}. For CS, neural networks were combined with the alternating direction method of multipliers~(ADMM) for efficient MRI reconstruction \cite{yang2018admm}. \cite{borgerding2017amp} learn the sparse linear inverse problem from the perspective of approximate message passing~(AMP) and follow-up work showed intensive studies~\cite{zhang2020amp} on image CS. The ISTA-Net+~\cite{zhang2018ista} focuses on modelling the iterative shrinkage thresholding algorithm~(ISTA) with neural networks, whereas the OPINE-Net~\cite{zhang2020optimization} obtained satisfactory results using a binary trainable sampling matrix. While the other models were trained with a fixed CS ratio, the ISTA-Net++~\cite{you2021ista} model trains at multiple CS ratios, reducing computational cost. Finally, COAST~\cite{you2021coast} is able to handle arbitrary sampling matrices and achieves promising results.




\subsection{Low-Rank Representation}

Low-rank representations characterize high-dimensional data with fewer vectors and effectively reveal the overall data structure~\cite{liu2010robust, kong2021infrared}. They are widely applied in the fields such as image restoration~\cite{gu2014weighted, han2021fine}, background modelling~\cite{wei2017self}, infrared small target detection~\cite{zhang2021infrared, zhang2019infrared, wang2021infrared}, and image compressive sensing~\cite{ravishankar2017low}. 
Whereas, due to the non-convex nature, its optimal convex approximation is widely adopted, i.e. the nuclear norm $\left\| \textbf{x} \right\|_* = \sum_i \sigma_i$, where $\sigma$ is the singular value. 
Even though low-rank representations helped model-driven approaches to achieve great sucess, the necessity of the singular value decomposition~(SVD) greatly limits computational efficiency. The SVD also complicates the integration of the low-rankness condition into a neural network.
\cite{cai2021learned} consider the low-rank term to be the product of two sub-matrices. Inspired by the tensor CP (CANDECOMP/PARAFAC) decomposition, Chen \textit{et.al}~\cite{chen2020tensor} treats a tensor of rank $r$ as the sum of multiple rank unit tensors and apply it as attention to semantic segmentation. Although this problem has been partially studied, it remains a challenging task to learn low-rank priors more efficiently and mapping them into DUNs reasonably.

\section{Proposed LR-CSNet}


In this section, we first define the low rank constrained CS problem and then present our specific update process in terms of optimization. 
Thereafter, we describe the process of mapping the optimization into neural networks and give details on the LR-CSNet. Finally, the training parameters and loss function are described in general.

We use the following notation. Plain font $\rho$ indicates scalars, bold lowercase $\textbf{x}$ indicates matrices and vectors, bold capital $\textbf{F}$ indicates deep features, and calligraphic font $\mathcal{G}$ indicates functions.

\subsection{Problem Formulation}

We constrain the L2 norm of the low-rank component $\textbf{l}$ and signal $\textbf{x}$, rather than using the nuclear norm directly, in order to circumvent the costly SVD, i.e. $\mathcal{R}(\textbf{x}) = \frac{1}{2} \left\| \textbf{x} - \textbf{l} \right\|^2_{\text{F}}$. 
Then we perform variable splitting to reduce the complex operations during optimization, specifically, we introduce an auxiliary variable $\textbf{z}$ as follows:

\begin{align}
    \begin{aligned}
    \min \limits_{\textbf{x}, \textbf{z}} & \left[\frac{1}{2} \left\| \boldsymbol{\Phi} \textbf{z} - \textbf{y} \right\|^2_{\text{F}} + \lambda \mathcal{G}(\textbf{x}) + \frac{\mu}{2} \left\| \textbf{z} - \textbf{l} \right\|^2_{\text{F}}\right] \text{s.t.} \quad \textbf{x} = \textbf{z}
    \label{eq:objectivefunction}
    \end{aligned}
\end{align}
Subsequently, we optimize the unconstrained cost function in Eq. \eqref{eq:costfunction}, where $\mu$ and $\beta$ are the penalty parameters

\begin{equation}
    \mathcal{L} \left( \textbf{x}, \textbf{z} \right) =  \frac{1}{2} \left\| \boldsymbol{\Phi} \textbf{z} - \textbf{y} \right\|^2_{\text{F}} + \lambda \mathcal{G}(\textbf{x}) + \frac{\mu}{2} \left\| \textbf{z} - \textbf{l} \right\|^2_{\text{F}} + \frac{\beta}{2} \left\| \textbf{x} - \textbf{z} \right\|^2_{\text{F}}
    \label{eq:costfunction}
\end{equation}

For a differentiable function $f(\textbf{x})$ with $\nabla f(\textbf{x})$ being $l$-Lipschitz continuous (i.e. $\forall \textbf{x}_1, \textbf{x}_2: \left\| \nabla f(\textbf{x}_1) - \nabla f(\textbf{x}_2) \right\|$ $\leq l \left\| \textbf{x}_1 - \textbf{x}_2 \right\|$, where $l$ is a constant), the Taylor expansion at $\textbf{x}_0$ leads to an upper bound 
$f(\textbf{x}) \le \hat{f}(\textbf{x}, \textbf{x}_0) =
f(\textbf{x}_0) + \left< \nabla f(\textbf{x}_0), \textbf{x} - \textbf{x}_0 \right> + \frac{l}{2} \left\| \textbf{x} - \textbf{x}_0 \right\|^2 = \frac{l}{2} \left\| \textbf{x} - \textbf{x}_0 + \frac{1}{l} \nabla f \left( \textbf{x}_0 \right) \right\|^2_2 + C$, where $C = - \frac{1}{2l} \left\| \nabla f(\textbf{x}_0) \right\|^2 + f(\textbf{x}_0)$.

Further, we optimize the variables $\textbf{z}$ and $\textbf{x}$ separately. The low-rank component $\textbf{l}^k$ is hypothesised to be independent of $\textbf{z}$, $\textbf{x}$ and is generated by LRGM.

\paragraph{Updating $\textbf{z}^k$:} The optimization objective is given by  
\begin{multline}
    \textbf{z}^k = \arg \min \limits_{\textbf{z}} \bigg[\frac{1}{2} \left\| \boldsymbol{\Phi} \textbf{z} - \textbf{y} \right\|^2_{\text{F}} + \\ \frac{\mu}{2} \left\| \textbf{z} - \textbf{l}^{k} \right\|^2_{\text{F}} + \frac{\beta}{2} \left\| \textbf{x}^{k-1} - \textbf{z} \right\|^2_{\text{F}}\bigg]\enspace.
    \label{eq:optimizez}
\end{multline}
In order to avoid the complex operations such as matrix inversion that occur in the update process, we perform the Taylor expansion at $\textbf{z}^{k-1}$ for the first term in Eq.~\eqref{eq:optimizez}, i.e. we replace $\frac{1}{2} \left\| \boldsymbol{\Phi} \textbf{z} - \textbf{y} \right\|^2_{\text{F}}$ by  $\frac{l_1}{4} \left\| \textbf{z} - \textbf{z}^{k-1} + \frac{1}{l_1} \boldsymbol{\Phi}^{\top} \left( \boldsymbol{\Phi} \textbf{z}^{k-1} - \textbf{y} \right) \right\| + C_1$, which as a linear function is $l_1$-Lipschitz continuous and get the update step using $s = l_1 + 2\mu + 2\beta$: 
\begin{equation}
    \textbf{z}^k = \frac{1}{s} \left( 2 \beta \textbf{x}^{k-1} + l_1 \textbf{z}^{k-1} + 2 \mu \textbf{l}^{k} - \boldsymbol{\Phi}^\top \boldsymbol{\Phi} \textbf{z}^{k-1} + \boldsymbol{\Phi}^\top \textbf{y} \right)
    \label{eq:updatez}
\end{equation}

\paragraph{Updating $\textbf{x}^k$:} The optimization objective is given by
\begin{equation}
    \textbf{x}^k = \arg \min \limits_{\textbf{x}} \left[\lambda \mathcal{G}(\textbf{x}) + \frac{\beta}{2} \left\| \textbf{x} - \textbf{z}^{k} \right\|^2_{\text{F}}\right]\enspace.
    \label{eq:optimizex}
\end{equation}

As a function of enforcing the signal to be sparse in some transform domain, $\mathcal{G} (\textbf{x})$ is not specifically mathematically constrained. Similarly, we perform a Taylor expansion of $\mathcal{G} (\textbf{x})$ at $\textbf{x}^{k-1}$, which is converted into the $\nabla \mathcal{G} (\textbf{x})$ form with L2 norm constraints and arrive at
\begin{equation}
    \textbf{x}^k = \frac{\lambda l_2}{\lambda l_2 + \beta} \textbf{x}^{k-1} +  \frac{\beta}{\lambda l_2 + \beta} \textbf{z}^{k} - \frac{\lambda}{\lambda l_2 + \beta} \nabla \mathcal{G} \left( \textbf{x}^{k-1} \right)\enspace.
    \label{eq:updatex}
\end{equation}
We replace the unknown function $\nabla \mathcal{G} (\textbf{x})$ with convolutional layers in LR-CSNet, which also satisfies that $\nabla \mathcal{G} (\textbf{x})$ is $l_2$-Lipschitz continuous.

\paragraph{Overall:} In end-to-end learning, we can set complex penalty parameters as learnable variables, so the overall optimization steps are 
\begin{align}
    \begin{aligned}
    \textbf{z}^k = & \rho^k_1 \textbf{x}^{k-1} +  \rho^k_2 \textbf{z}^{k-1} + \left( 1 - \rho^k_1 - \rho^k_2 \right) \textbf{l}^{k} \\
    & - \eta^k \boldsymbol{\Phi}^\top \boldsymbol{\Phi} \textbf{z}^{k-1} + \eta^k \boldsymbol{\Phi}^\top \textbf{y} \\
    \textbf{x}^k = & \alpha^k \textbf{x}^{k-1} + \left( 1 - \alpha^k \right) \textbf{z}^{k} - \gamma^k \nabla \mathcal{G} \left( \textbf{x}^{k-1} \right) \enspace,
    \label{eq:updateprocess}
    \end{aligned}
\end{align}
where $\rho_1 = \frac{2 \beta}{l_1 + 2\mu + 2\beta}$, $\rho_2 = \frac{l_1}{l_1 + 2\mu + 2\beta}$, $\eta = \frac{1}{l_1 + 2\mu + 2\beta}$, $\alpha = \frac{\lambda l_2}{\lambda l_2 + \beta}$ and $\gamma = \frac{\lambda}{\lambda l_2 + \beta}$. These parameters are trained independently in each reconstruction stage.



\subsection{Network Architecture}
\label{subsect:networkarch}

In this section we elaborate on the network architecture and module design of LR-CSNet based on the optimization process of Eq.~\eqref{eq:updateprocess}. 
As shown in Fig.~\ref{fig:networkarch}, given an original signal $\originalsignal \in \mathbb{R}^{\sqrt{N}\times \sqrt{N}}$, we perform sampling and end-to-end image reconstruction through the network.

During sampling, the original image $\originalsignal$ passes through a convolutional layer with a kernel size and step size of 33, where the input and output channels are $1$ and $M$, respectively. In this way, the sampling process for $\textbf{y} = \boldsymbol{\Phi} \originalsignal$ is simulated and the observation $\textbf{y} \in \mathbb{R}^{1\times 1\times M}$ is obtained. 

In the reconstruction phase, $\textbf{y}$ is passed through convolutional layers with kernel size and stride of $1$, where the input and output channels are $M$ and $33\times 33$ respectively. This operation is used to simulate $\textbf{x}^0 = \boldsymbol{\Phi}^{\top} \textbf{y}$, where the convolutional layer share weights with the one in the sampling process. Then the reconstructed signal is reshaped to $\textbf{x}^0 \in \mathbb{R}^{\sqrt{N}\times \sqrt{N}}$ by $\textit{PixelShuffle}(33)$~\cite{zhang2020optimization}. The reconstructed image is then passed through $K$ reconstruction stages to simulate the iterative updates. Each reconstruction stage consists of two modules: LRGM and GDPM. 

\subsubsection{Low-Rank Generation Module (LRGM)} LRGM is used to generate the low-rank matrix $\textbf{l}^k$ of the current stage, which contains the majority of the information in the background. 
A low-rank matrix can be considered as the result of multiplying two sub-matrices together, i.e. $\textbf{l} = \textbf{p} \textbf{q}$, where $\textbf{p} \in \mathbb{R}^{\sqrt{N}\times r}$, $\textbf{q} \in \mathbb{R}^{r\times \sqrt{N}}$, and $r$ is the rank number. 
LRGM takes the updated variables from the previous stage as input and concatenates it with the transferred tensor $\textbf{F}^{k-1}$ after one convolutional layer, as shown in Fig. \ref{fig:networkarch}.
Subsequently, the deep feature is adaptively pooled into two tensors of scale $\sqrt{N} \times r \times C$ and $r \times \sqrt{N} \times C$ according to the rank number $r$ respectively, where $C$ is the channel number. 
The two sub-matrices $\textbf{p}$ and $\textbf{q}$ are obtained through two 1$\times$1 convolutional layers for feature separation and dimensionality reduction.
Finally, these sub-matrices are multiplied to obtain the updated low-rank matrix $\textbf{l}^k$. 
In this way LRGM is able to guarantee that $\operatorname{rank}(\textbf{l}^k) \le r$.

\subsubsection{Gradient Descent and Proximal Mapping (GDPM)} GDPM is used to update the variables $\textbf{z}^k$ and $\textbf{x}^k$ according to Eq. \eqref{eq:updateprocess}, whereby the scalars are learnable variables. 
In Fig. \ref{fig:networkarch}, after obtaining $\textbf{z}^k$, it is concatenated with the low-rank matrix $\textbf{l}^k$ and passed through a convolution layer. 
$\textbf{l}^k$ contains more structured image information and can provide guidance for the neural network in learning image details. Then we simulate the function $\nabla \mathcal{G}$ with two dense blocks (DBs)~\cite{zhang2018residual}. 
Since $\nabla \mathcal{G}$ is learning high-frequency details in the image, the residual connections by \cite{zhang2018residual} were not applied here. It is worth mentioning that DB is essentially an accumulation of multiple convolutional layers, which clearly satisfies Lipschitz continuous, guaranteeing the validity of this module. In addition, the last deep feature is concatenated with $\textbf{F}^{k-1}$ and the transferred tensor $\textbf{F}^{k}$ is updated through a 1x1 convolutional layer. Finally, the transferred tensor $\textbf{F}^{k}$, the updated variables $\textbf{z}^{k}$, and $\textbf{x}^{k}$ are delivered to the next reconstruction stage.


\subsection{Network Parameter and Loss Function}

The trainable parameters in LR-CSNet consist of four components: 1) the same measurement matrix $\boldsymbol{\Phi}$ in each reconstruction stage, 2) the auxiliary scalars $\rho_1$, $\rho_2$, $\eta$, $\alpha$, $\gamma$, 3) the network weights $\Theta_{l}$ in LRGM, and 4) the weights $\Theta_{g}$ in GDPM.
Thus, all training parameters are denoted as $\Theta = \{ \boldsymbol{\Phi} \} \cup \{ \rho^k_1, \rho^k_2, \eta^k, \alpha^k, \gamma^k \}^{K}_{k=1} \cup \{ \Theta^k_{l}, \Theta^k_{g} \}^{K}_{k=1}$, where $K$ is the total reconstruction stages.
$\boldsymbol{\Phi}$ and $\boldsymbol{\Phi}^{\top}$ share  weights~\cite{zhang2020optimization}.

The loss function of the network, as it is common practice~\cite{zhang2020optimization}, consists of two components for the given training data $\{ \originalsignal_i \}^{\trainnumber}_{i=1}$: the fidelity loss $\mathcal{L}_{\text{fidelity}}$ to ensure that the reconstruction result $\textbf{x}^K_i$ closely approximates the input $\originalsignal_i$ and the orthogonal loss $\mathcal{L}_{\text{orth}}$ to impose an orthogonal constraint on the measurement matrix. The combined loss function is
\begin{align}
    \begin{aligned}
    \mathcal{L} (\Theta) & = \mathcal{L}_{\text{fidelity}} + \tau \mathcal{L}_{\text{orth}} \\
    & = \frac{1}{N\trainnumber} \sum^{\trainnumber}_{i=1} \left\| \textbf{x}^K_i - \originalsignal_i \right\|^2_{\text{F}} + \frac{\tau}{M^2} \left\| \boldsymbol{\Phi} \boldsymbol{\Phi}^{\top} - \textbf{E} \right\|^2_{\text{F}}\enspace,
    \label{eq:loss}
    \end{aligned}
\end{align}
where $\trainnumber$ is the total amount of training data, \textbf{E} the unit matrix, and $\tau$ a constant (set to $0.01$ for our experimental evaluation).

\begin{table}
    \renewcommand\arraystretch{1.3}
    \begin{center}
    \caption{Ablation study on the effects of our introduced model components with a fixed CS ratio of 25\%. 
    \label{Tab:Ablationmodules}}
        \begin{tabular}{ccc|c|c}
        \Xhline{1.3pt}
        \multirow{2}{*}{LRGM} & \multirow{2}{*}{Dense} & \multirow{2}{*}{Trans} & \multicolumn{2}{c}{PSNR/SSIM} \\
        \cline{4-5}
        & & & Set11 & BSD68 \\
        \hline
        $\surd$ & - & - & 35.12/0.9536 & 31.87/0.9127 \\
        - & $\surd$ & - & 35.27/0.9552 & 31.94/0.9136 \\
        - & - & $\surd$ & 35.28/0.9547 & 31.95/0.9141 \\
        \hline
        - & $\surd$ & $\surd$ & 35.47/0.9565 & 32.09/0.9158 \\
        $\surd$ & - & $\surd$ & 35.37/0.9556 & 32.02/0.9148 \\
        $\surd$ & $\surd$ & - & 35.44/0.9561 & 32.00/0.9148 \\
        
        \CC{90} $\surd$ & \CC{90} $\surd$ & \CC{90} $\surd$  & \CC{90} \textbf{35.54/0.9567} & \CC{90} \textbf{32.12/0.9162} \\
        \Xhline{1.3pt}
        \end{tabular}
    \end{center}
    
\end{table}

\section{Experiments}

In this section we first give details on the widely applied datasets, the evaluation metrics, and the network implementation.Then, we demonstrate the validity of each module in this paper through extensive ablation studies and investigate the effect of key parameters.  Finally, we compare LR-CSNet with other state-of-the-art methods in both quantitative and qualitative aspects to validate the performance of our approach.

\begin{figure}
  \centering
  \includegraphics[width=0.43\textwidth]{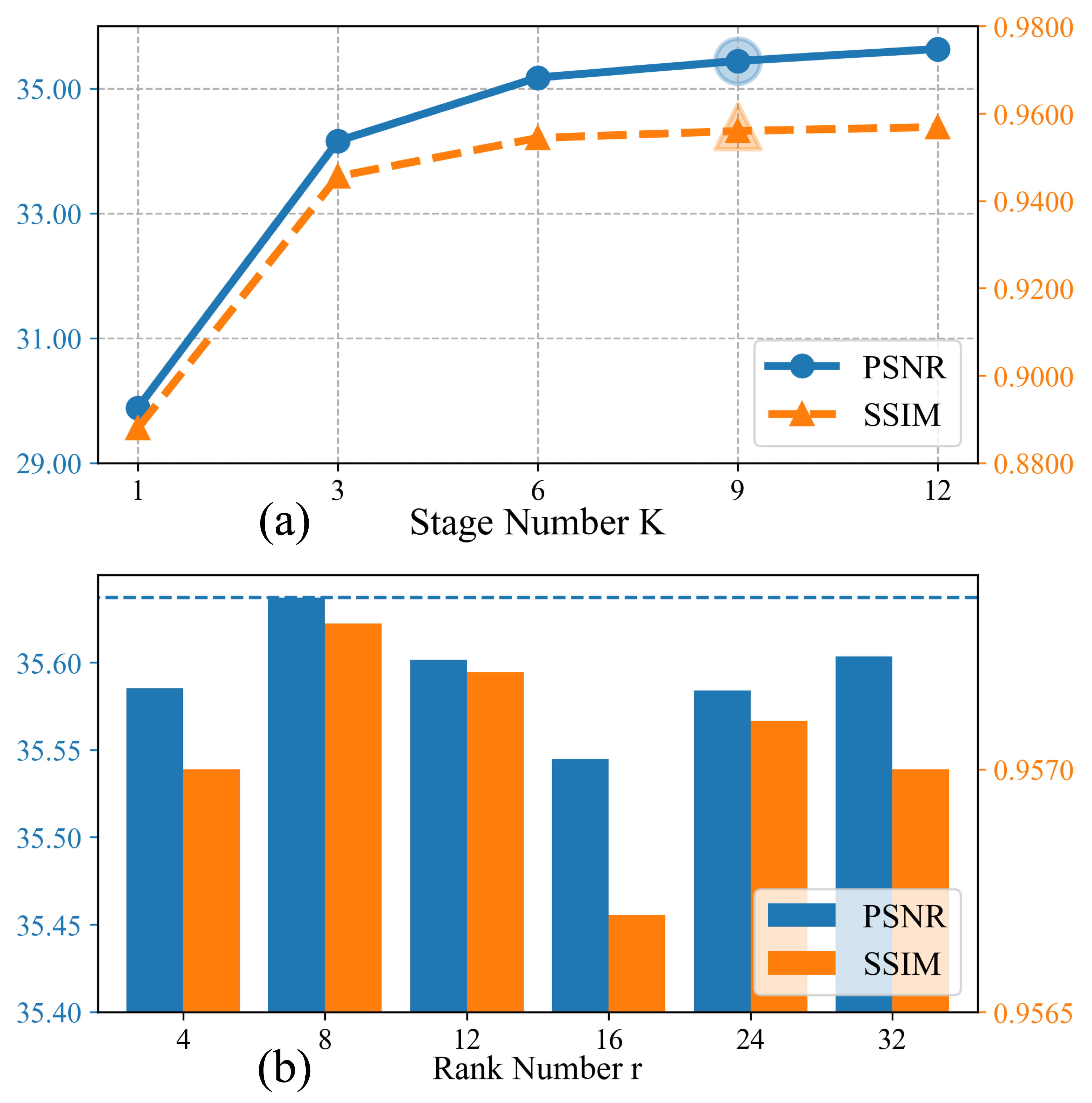}
  \caption{Experiments the number of reconstruction stages $K$ and the rank number $r$ in LRGM with CS of 25\% on Set11. We create a trade-off between performance and computing efficiency by setting $K=9$ and $r=8$ (see increased marker size for $K$ and the dashed line for $r$).}
  \label{fig:stagerank}
\end{figure}

\begin{table*}
    \renewcommand\arraystretch{1.3}
    \begin{center}
        \caption{Quantitative comparison of average PSNR/SSIM for different CS ratios on the Set11, BSD68, and Urban100. 
        \label{table:comparesota}}

        \begin{tabular}{c|c|cccccc}
        
        \Xhline{1.3pt}
        Dataset & Ratio & ISTA-Net+ & CSNet+ & AdapRecon & OPINE-Net+ & AMP-Net & LR-CSNet \\
        
        \Xhline{1.3pt}

        \multirow{5}{*}{Set11} & 1\% & 17.42/0.4029 & 19.87/0.4977 & 19.63/0.4848 & 20.02/0.5362 & 20.04/0.5132 & \CC{90} \textbf{20.85/0.5583} \\
        & 4\%  & 21.32/0.6037 & 23.93/0.7338 & 23.87/0.7279 & 25.69/0.7920 & 24.64/0.7527 & \CC{90} \textbf{26.16/0.8040} \\
        & 10\% & 26.64/0.8087 & 26.04/0.7971 & 27.39/0.8521 & 29.81/0.8884 & 28.84/0.8765 & \CC{90} \textbf{30.35/0.8987} \\
        & 25\% & 32.59/0.9254 & 29.98/0.8932 & 31.75/0.9257 & 34.86/0.9509 & 34.42/0.9513 & \CC{90} \textbf{35.64/0.9573} \\
        & 50\% & 38.11/0.9707 & 34.61/0.9435 & 35.87/0.9625 & 40.17/0.9797 & 40.12/0.9818 & \CC{90} \textbf{41.03/0.9826}  \\
        
        \hline
        \multirow{5}{*}{BSD68} & 1\% & 19.14/0.4158 & 21.91/0.4958 & 21.50/0.4825 & 21.88/0.5162 & 21.97/0.5086 & \CC{90} \textbf{22.32/0.5282} \\
        & 4\%  & 22.17/0.5486 & 24.63/0.6564 & 24.30/0.6491 & 25.20/0.6825 & 25.40/0.6985 & \CC{90} \textbf{25.53/0.6972} \\
        & 10\% & 25.32/0.7022 & 27.02/0.7864 & 26.72/0.7821 & 27.82/0.8045 & 27.41/0.8036 & \CC{90} \textbf{28.21/0.8159} \\
        & 25\% & 29.36/0.8525 & 30.22/0.8918 & 30.10/0.8901 & 31.51/0.9061 & 31.56/0.9121 & \CC{90} \textbf{32.12/0.9162} \\
        & 50\% & 34.04/0.9424 & 34.82/0.9590 & 33.60/0.9479 & 36.35/0.9660 & 36.64/0.9707 & \CC{90} \textbf{37.29/0.9720} \\
        
        \hline
        \multirow{5}{*}{Urban100} & 1\% & 16.90/0.3846 & 19.26/0.4632 & 19.14/0.4510 & 19.38/0.4872 & 19.62/0.4967 & \CC{90} \textbf{19.65/0.4971} \\
        & 4\%  & 19.83/0.5377 & 21.96/0.6430 & 21.92/0.6390 & 23.36/0.7114 & 22.82/0.6963 & \CC{90} \textbf{23.41/0.7210} \\
        & 10\% & 24.04/0.7378 & 24.76/0.7899 & 24.55/0.7801 & 26.93/0.8397 & 26.05/0.8287 & \CC{90} \textbf{27.41/0.8547} \\
        & 25\% & 29.78/0.8954 & 28.13/0.8827 & 28.21/0.8841 & 31.86/0.9308 & 30.94/0.9273 & \CC{90} \textbf{32.50/0.9391} \\
        & 50\% & 35.24/0.9614 & 32.97/0.9503 & 31.88/0.9434 & 37.23/0.9741 & 36.54/0.9744 & \CC{90} \textbf{37.87/0.9776} \\

        \Xhline{1.3pt}
        \end{tabular}
    \end{center}

\end{table*}

\subsection{Datasets and Evaluation Metrics}
We test LR-CSNet on three natural image dataset benchmarks that are widely used in CS: Set11~\cite{Kulkarni_2016_CVPR}, BSD68~\cite{martin2001database}, and Urban100~\cite{huang2015single}. As training data, we use image patches $\{ \originalsignal_i \}^{\trainnumber}_{i=1}$ of size $33 \times 33$ as published in~\cite{zhang2018ista}, where the total number is $\trainnumber = 88912$. For fine-tuning, we train with an additional $36000$ image patches of size $99 \times 99$ from BSD300~\cite{martin2001database}
, which is also publicly available. 
As evaluation metrics, we choose peak signal-to-noise ratio (PSNR) and the structural similarity index measure (SSIM), which are widely adopted in image restoration, with higher values of both indicating better reconstruction results.

\subsection{Implementation Details}
Our implementation is based on PyTorch~\cite{paszke2019pytorch} and all experiments are performed on NVIDIA Titan RTX. We train the network on a set of CS ratios $\{1\%, 4\%, 10\%, 25\%, 50\%\}$, where we train 150 epochs using $33 \times 33$ image patches with batch size of 128, followed by a fine-tuning phase of 100 epochs using $99 \times 99$ image patches with batch size of 32. We optimize the parameters using Adam~\cite{kingma2014adam} with a momentum of~$0.9$ and weight decay of~$0.999$. The learning rate was set to a constant $10^{-4}$.

\begin{figure}[t]
  \centering
  \includegraphics[width=0.49\textwidth]{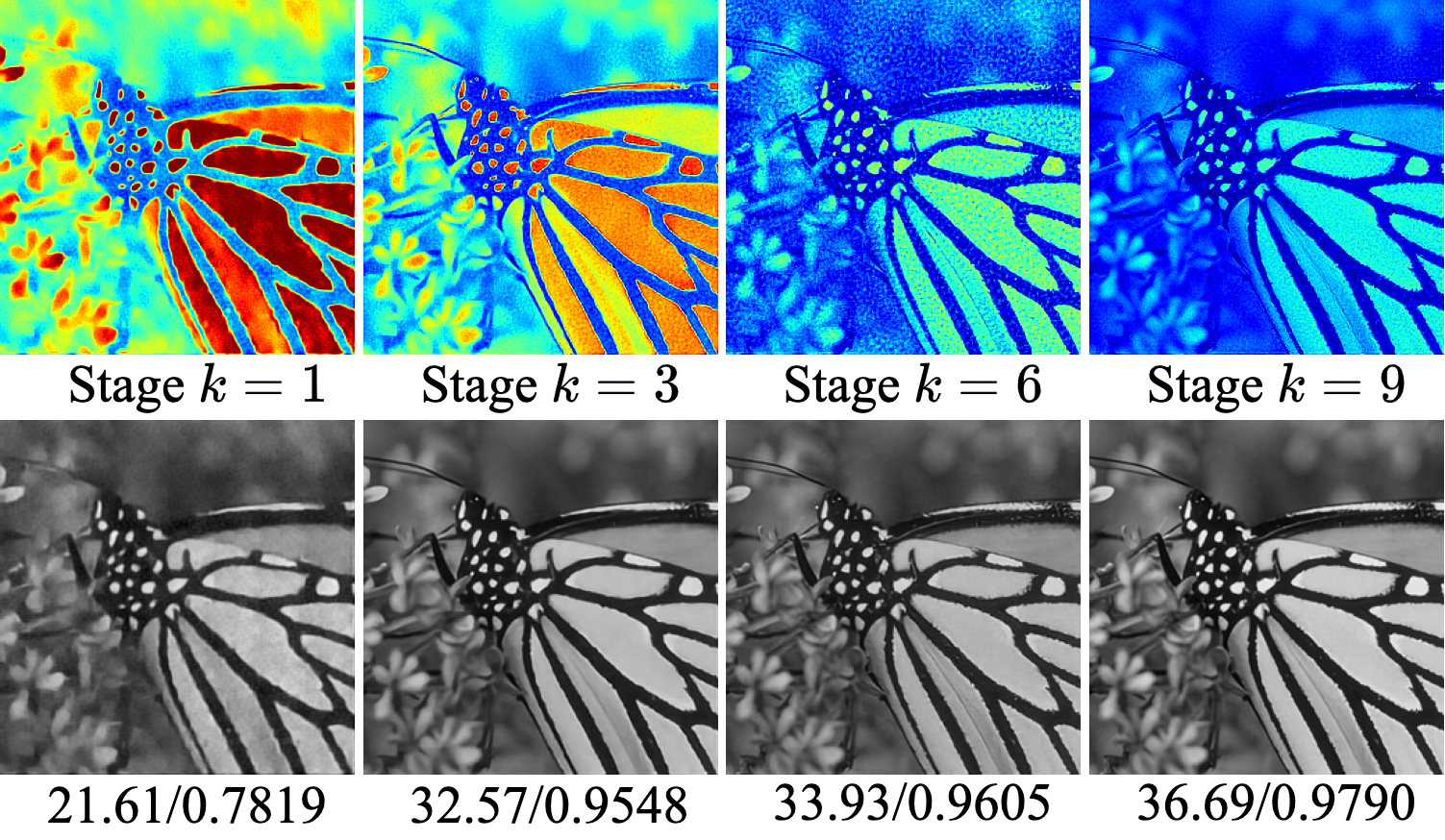}
  \caption{Illustration of intermediate results on Set11 'Monarch' with CS ratio of 25\%. The feature map is the output of $\nabla \mathcal{G}^k$ (upper) and the reconstruction result $\textbf{x}^k$. The metrics of the corresponding stage (lower) are shown respectively, where $k \in \{1, 3, 6, 9\}$.}
  \label{fig:stageresults}
\end{figure}

\begin{figure*}[ht]
  \centering
  \includegraphics[width=0.85\textwidth]{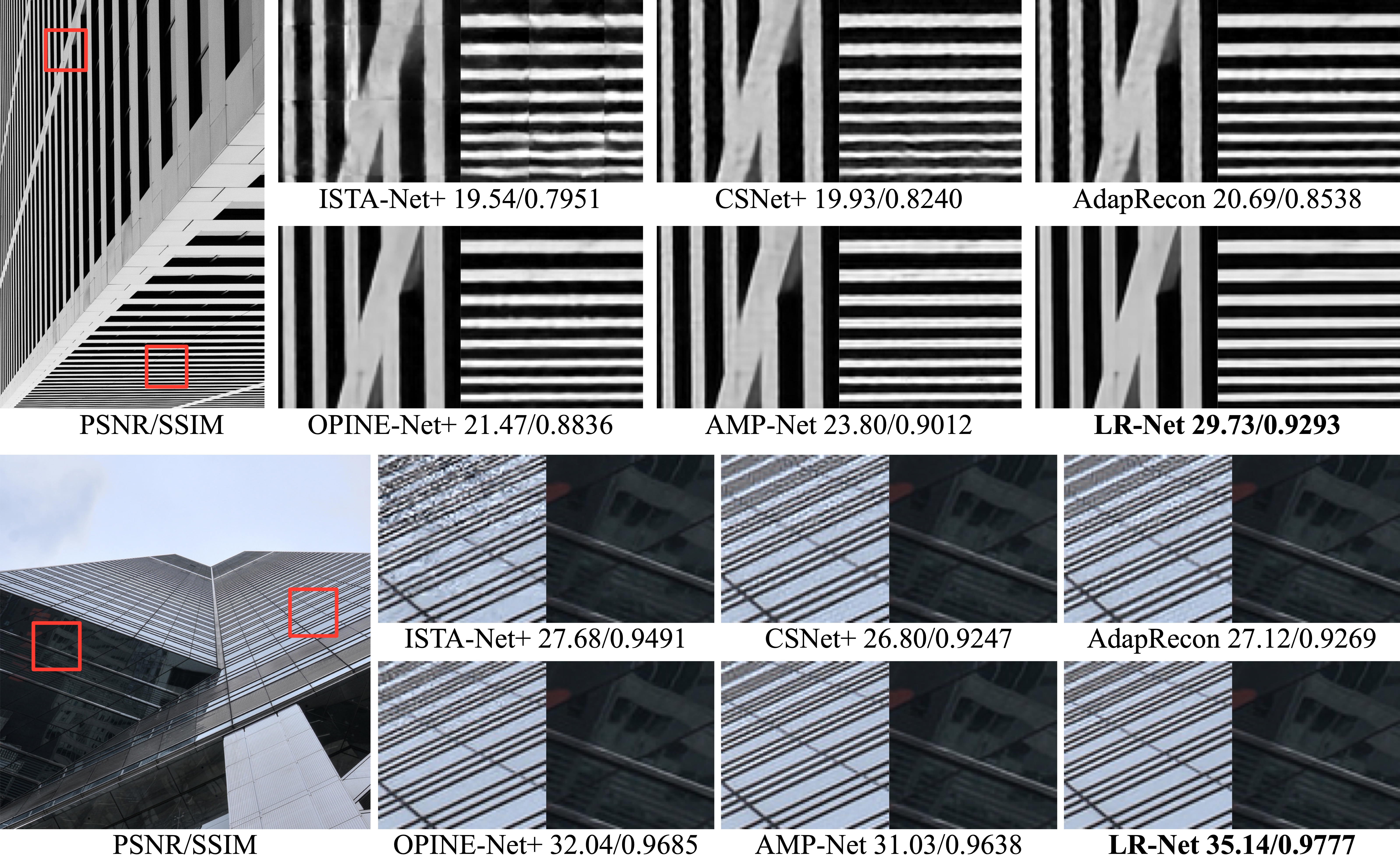}
  \caption{Visual comparison on 'img\_011' with CS ratio of 10\% (upper) and on 'img\_059' with CS ratio of 50\% (lower). The best performance is highlighted.}
  \label{fig:qualitative}
\end{figure*}

\subsection{Ablation Study and Parameter Setting}

We present the ablation study in Fig.~\ref{fig:stagerank}~\ref{fig:stageresults}, and Table~\ref{Tab:Ablationmodules} to explore the impact of each module and the changes in key parameters.

\subsubsection{Impact of LRGM} Table~\ref{Tab:Ablationmodules} explores the effectiveness of each module by comparing each possible combination. Removing 'LRGM' means removing the constraint on low-rank from the problem formulation in Eq.~\eqref{eq:objectivefunction}, with the derivation and settings remaining the same as before. The results show that LRGM always contributes to the performance.

\subsubsection{Impact of Dense} Removing 'Dense' is to replace the dense block in Fig.~\ref{fig:networkarch} with 6 convolutional layers, which implies a reduction in network parameter number and feature reuse capability. Again, the results demonstrate a degradation in performance without 'Dense'.

\subsubsection{Impact of Transmission} Transmission is the integration of deep features from the previous reconstruction stages into the current stage, which theoretically enables a more effective aggregation of information. Removing 'Trans' is removing the $\textbf{F}$ in Fig.~\ref{fig:networkarch}. 
The experiments show that 'Trans' does improve the reconstruction accuracy of the network.

\subsubsection{Rank Number $r$} From Table~\ref{fig:stagerank}, we employ ranks $r \in \{4,8,12,16,24,32\}$ for the LRGM module to analyze their impact. 
In general, a larger rank number indicates that more information can potentially be learned. 
However, Fig.~\ref{fig:stagerank} shows that $r = 8$ performs best. 
This indicates that larger ranks lead to redundant information that does not help improve network performance. We therefore set $r = 8$.


\subsubsection{Stage Number $K$} We explore the performance gain from the number of reconstruction stages $K$. As shown in Fig.~\ref{fig:stagerank}, we set $K \in \{ 1,3,6,9,12 \}$. The process of making $K$ larger brings a significant gain and also increases the number of network parameters. We found a trade-off at $K = 9$.

\subsection{Comparison With State-of-the-Art}

\subsubsection{Quantitative Evaluation} We compare LR-CSNet with five state-of-the-art methods, including two DNUNs: CSNet+~\cite{shi2019image}, AdapRecon~\cite{lohit2018convolutional}, and three DUNs: ISTA-Net+~\cite{zhang2018ista}, OPINE-Net+~\cite{zhang2020optimization}, and AMP-Net~\cite{zhang2020amp}. 
We summarize the evaluation metrics of these methods on multiple datasets in Table \ref{table:comparesota}. 
It can be seen that deep non-folding networks are stacking more convolutional layers which does not increase performance. Whereas ISTA-Net+ operates directly on deep features to simulate soft-thresholding, which limits the representation capability, resulting in poor performance. Meanwhile, OPINE-Net+ uses convolutional layers to simulate the analytical solution of an optimization problem such as the sum of L2 norm and L1 norm. AMP-Net focuses on removing the boundary effects between image patches using denoising techniques. These approaches ignore the low-rank properties of the image patches, resulting in networks that capture structural information only to a limited extent. 
As shown in the table, LR-CSNet achieves the best reconstruction results at multiple CS ratios.


\subsubsection{Qualitative Evaluation} 
Fig.~\ref{fig:stageresults} visualizes the reconstruction results at each stage, where higher stages are reconstructed more acculately and the information learnt by $\nabla \mathcal{G}^k$ becomes increasingly more detailed. 
In addition, to illustrate the reconstruction effect of LR-CSNet more intuitively, we show the reconstruction effect of state-of-the-art approaches and LR-CSNet on two images as in Fig.~\ref{fig:qualitative}, where the red-boxed parts are enlarged and placed on the right side. The corresponding method and evaluation metrics are listed below and the best value is highlighted. Compared to the other methods, LR-CSNet is better at capturing the overall structure of the image and retains detailed information. This is due to the network taking into account the low-rank attributes of the image patches, together with $\nabla \mathcal{G}$ to learn high-frequency information, leading to its ability to obtain better reconstruction accuracy.

\section{Conclusion}

In this paper, we propose a deep unfolding network for natural image compressive sensing (CS) called LR-CSNet.
As real-world image patches are often well-
represented by low-rank approximations, we add a low-rank prior to the CS reconstruction.
We unfold the corresponding iterative optimization problem using variable splitting, leading to a neural network for CS that can be trained end-to-end.  
Extensive experiments support the effectiveness of our approach.

\bibliographystyle{plain}
\bibliography{ref}

\end{document}